\pdfoutput=1
\documentclass[11pt]{article}

\usepackage{ACL2023}

\usepackage{times}
\usepackage{latexsym}

\usepackage[T1]{fontenc}
\usepackage[utf8]{inputenc}

\usepackage{microtype}

\usepackage{inconsolata}
\usepackage{graphicx}
\usepackage{array}
\usepackage{algorithm}
\usepackage{algorithmic}
\usepackage{amsmath}
\usepackage{cleveref}
\usepackage{multirow}
\usepackage{booktabs}
\usepackage{makecell}
\usepackage{CJKutf8}
\usepackage{colortbl}   
\title{Enhancing User Engagement in Socially-Driven Dialogue through Interactive LLM Alignments}
\author{Jiashuo Wang$^{1}$\quad Kaitao Song$^{2}$\quad Chunpu Xu$^{1}$\quad Changhe Song$^{1}$\quad Yang Xiao$^{1}$\\\textbf{Dongsheng Li}$^{2}$\quad \textbf{Lili Qiu}$^{2}$\quad \textbf{Wenjie Li}$^{1}$ \\
$^{1}$Department of Computing, The Hong Kong Polytechnic University \quad $^{2}$Microsoft Research Asia  \\
\small{\texttt{\{csjwang,cswjli\}@comp.polyu.edu.hk} \quad \texttt{\{yang-alan.xiao,changhe.song,chun-pu.xu\}}@connect.polyu.hk}\\
\small{\texttt{\{kaitaosong\}@microsoft.com}}}

\begin{document}
\maketitle
\begin{abstract}
Enhancing user engagement through interactions plays an essential role in socially-driven dialogues. While prior works have optimized models to reason over relevant knowledge or plan a dialogue act flow, the relationship between user engagement and knowledge or dialogue acts is subtle and does not guarantee user engagement in socially-driven dialogues. To this end, we enable interactive LLMs to learn user engagement by leveraging signals from the future development of conversations. Specifically, we adopt a more direct and relevant indicator of user engagement, i.e., the user's reaction related to dialogue intention after the interaction, as a reward to align interactive LLMs. To achieve this, we develop a user simulator to interact with target interactive LLMs and explore interactions between the user and the interactive LLM system via \textit{i$\times$MCTS} (\textit{M}onte \textit{C}arlo \textit{T}ree \textit{S}earch for \textit{i}nteraction). In this way, we collect a dataset containing pairs of higher and lower-quality experiences using \textit{i$\times$MCTS}, and align interactive LLMs for high-level user engagement by direct preference optimization (DPO) accordingly. Experiments conducted on two socially-driven dialogue scenarios (emotional support conversations and persuasion for good) demonstrate that our method effectively enhances user engagement in interactive LLMs.
\end{abstract}

\section{Introduction}
Large language models (LLMs) have recently made remarkable advancements, enabling their application across diverse domains, such as hypothesis generation \cite{zhou2024hypothesis} and mental health counseling \cite{qiu2024interactive}. Many of these applications rely on interactive LLMs, where user engagement is essential in achieving various interaction goals \cite{liu2024compeer,furumai2024zero}. 

Engagement can take on different meanings and indicators depending on the context, including the intensity and duration of emotion, attention, and other forms of involvement \cite{fredricks2004school}. Conversation length can serve as a reasonable indicator in casual chitchat scenarios \cite{yu2017learning}, whereas emotional expression and problem recognition are likely more meaningful indicators in emotional support conversations \cite{chen2023soulchat,holdsworth2014client}. While engagement in chitchat has been explored in prior work \cite{yu2019gunrock,yu2017learning,yu2016wizard}, the engagement in socially-driven dialogues (emotional support and persuasion for good), has been less directly investigated.

Prior research on conversational language models has focused on enhancing either models' reasoning abilities over external knowledge \cite{deng2023knowledge,wang2022care} or their planning capabilities to design turn-level dialogue acts for smoother conversational flow \cite{yu2023prompt,liu-etal-2021-towards, wang2019persuasion}. These approaches are based on the logic that incorporating relevant knowledge and adopting appropriate dialogue acts can significantly increase the likelihood of engaging users. However, the relationship between user engagement and knowledge or dialogue acts is subtle rather than straightforward. For instance, while questions or informative content can encourage users to talk more and stay engaged, a poorly framed question or an overload of information may be off-putting \cite{chen2023soulchat}. In addition, the incorporation of external knowledge or dialogue acts may not be efficient in the era of LLMs. 

A more direct and relevant indicator of user engagement is the user's reaction that occurs after the interaction and is related to the dialogue's intended goal. Unlike dialogue acts or domain knowledge, which can often be identified within a single turn or exchange, user engagement is shaped by the cumulative effect of the entire conversation. Therefore, future development of conversations can benefit interactive LLMs in learning response generations towards better user engagement.

This work aims to leverage user's reaction related to the dialogue intention at the end of interactions to align interactive LLMs for enhanced user engagement. To achieve this, we collect interaction experiences between interactive LLMs and user simulators using a Monte Carlo Tree Search for interaction (\textit{i$\times$MCTS}). Next, the level of user engagement is detected using rule-based methods and regular expression matching. Finally, interactive LLMs are optimized via Direct Preference Optimization (DPO) \cite{rafailov2024direct} to increase the likelihood of generating responses that lead to stronger user engagement, while decreasing the likelihood of responses associated with weaker engagement.

For experiments, we select emotional support conversations and persuasion for good as evaluation tasks, as user engagement plays a crucial role in affecting user feelings, cognition and behaviors. When evaluating interactive LLMs optimized by the proposed method, we conduct both automated interactive evaluation and human subjective evaluation. Experimental results demonstrate that our method effectively enhances user engagement levels in interactions.

Our contributions can be summarized as follows:
\begin{itemize}
    \item This work focuses on enhancing user engagement in socially-driven dialogues, which has been less investigated directly in prior work.
    \item Interactive LLMs are trained to learn from future dialogue outcomes, derived by \textit{i$\times$MCTS}, via DPO to enhance user engagement.
    \item Experimental evaluations on emotional support conversations and persuasion for good demonstrate the effectiveness of our approach.
\end{itemize}
\section{Related Work}
\subsection{Engagement of Interactive LLMs}
Interactive LLMs play a crucial role in various applications, including emotional support \cite{qiu2024interactive, liu2024compeer} and persuasive dialogue \cite{furumai2024zero}. Ensuring user engagement is critical in these conversations, as it fosters deeper connections and a better understanding of users.
To enhance engagement, prior research has primarily focused on improving models' reasoning and planning capabilities. 
Existing approaches can be broadly categorized into two strategies: (1) reasoning enhancement, which incorporates external knowledge to improve response quality \cite{deng2023knowledge, wang2022care}, and (2) planning-based response generation, which designs turn-level response strategies to guide conversation flow \cite{yu2023prompt, liu-etal-2021-towards, wang2019persuasion}.
However, as LLMs scale up, these methods may become less effective and computationally inefficient. 
In this work, we take a more intuitive approach by leveraging user feedback in an interaction as a direct reward to align the LLM. 
Specifically, we use the user's final action or state at the end of the conversation as a criterion for measuring user engagement.
To achieve this, we enhance the model’s reasoning and planning capabilities using MCTS.

\subsection{MCTS with LLM}
% Monte Carlo Tree Search (MCTS) has recently emerged as a powerful technique for enhancing the reasoning and planning capabilities of LLMs, especially in mathematical, scientific, and commonsense reasoning tasks \cite{zhang2024rest,xie2024monte}. 
MCTS is a heuristic search algorithm that has been successfully applied to tasks requiring complex reasoning and planning, such as Atari \cite{ye2021mastering} and Go \cite{silver2016mastering}. It explores the decision space by iteratively constructing a search tree, balancing exploration and exploitation to identify optimal actions.
Recently, MCTS has been integrated with LLMs to enhance their reasoning and planning capabilities. 
It constructs a search tree by sampling model outputs based on the LLM’s policy.
Subsequently, paired outputs with higher and lower estimated rewards can be extracted and distilled into the LLM’s policy via supervised fine-tuning (SFT) or DPO \cite{rafailov2024direct}.
This approach has demonstrated success across various domains, including mathematical problem-solving \cite{wang2024towards,zhang2024rest,feng2023alphazero}, scientific reasoning \cite{xie2024monte}, and applications in programming and games \cite{zhou2023language}.
Beyond the above domains, MCTS has been applied to conversational tasks. Beyond these domains, MCTS has also been applied to conversational tasks. For instance, \citet{li2024planning} and \citet{yu2023prompt} utilize MCTS to enhance dialogue act planning, improving user-LLM interactions. Unlike these two approaches, we directly apply MCTS to optimize the interactive LLM itself, leading to greater effectiveness.

\section{Method}
We decide to align interactive LLMs for enhanced user engagement using DPO \cite{rafailov2024direct}. Thus, we construct a pairwise dataset, where each instance consists of a dialogue history and two candidate responses for the current turn. The two responses lead to higher and lower levels of user engagement, respectively. Accordingly, we propose \textit{i$\times$MCTS} (\textit{MCTS} for \textit{i}nteraction) involving a user simulator.

\subsection{User Simulator and Engagement}
\paragraph{User Simulator} We develop a user simulator to model a specific user's immediate reactions and final engagement level after a conversation. The user simulator is expected to interact like a user given a user condition $\mathcal{U}$, contents of which vary depending on the scenarios. $\mathcal{U}$ includes the user's troubles, feelings, and thoughts in the context of emotional support conversation, while encompasses psychological and demographic information in persuasion for good. With $\mathcal{U}$, the user simulator generates outputs at $i^{\text{th}}$ turn based on the history conversation $\mathcal{C}$, formulated as follows:
\begin{equation}
    \begin{aligned}
        o_i &= P_u(\mathcal{U}, \mathcal{C}_i), \\
        \mathcal{C}_i &= 
        \begin{aligned}
            &\begin{cases} 
                \emptyset, & \text{if } i = 0, \\ 
                \{u_0, a_1, \dots, a_{i-1}\}, & \text{otherwise}.
            \end{cases}
        \end{aligned}
    \end{aligned}
\end{equation}
$o_i$ represents the user simulator's emotional \& cognitive states and the response at the $i^{\text{th}}$ turn in emotional support conversations, whereas in persuasion for good, it denotes only the response at the $i^{\text{th}}$ turn. $\emptyset$ indicate ``'' (empty). Throughout the interaction, the user simulator is designed to behave like a user in datasets. Therefore, we use a simple and effective method--supervised finetuning (SFT)--to train the user simulators.

\paragraph{Engagement} We extract contents from the end of the conversation to determine the level of user engagement. In emotional support conversations, a full expression of feelings and thoughts is considered as a sign of user engagement, denoted by 1. Otherwise, the engagement level is denoted by 0. Therefore, we use regular expression matching to detect whether there is still any negative feelings and thoughts in the user simulator's emotional \& cognitive states in its output. In persuasion for good, we use the donation amount to reflect the user engagement, and use regular expression matching to extract the decided donation amount from the last user utterance. The engagement level is $\frac{\$\text{donation}}{2}$, where \$2 is the max donation amount. 

\subsection{MCTS for Interaction}\label{sec:mcts}
Unlike mathematical or scientific problems \cite{zhang2024rest, xie2024monte}, scenarios focused on user engagement inherently involve interactions. Additionally, user engagement can be assessed at the end of the interaction. Thus, when performing \textit{i$\times$MCTS}, we involve a user simulator.
Consequently, the tree comprises two types of nodes: user nodes and model nodes, with each model node always followed by a user node. Each node is an utterance. While user reactions can be diverse and could be modeled by generating multiple user utterances, we simplify the structure by considering only a single user node per model node.

\begin{figure}[htb]
\centering
\includegraphics{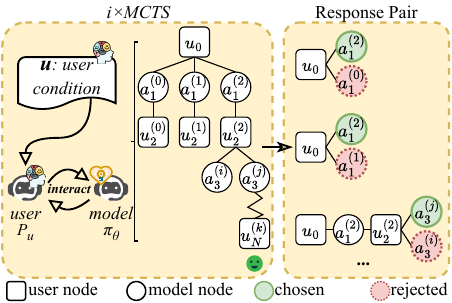} 
\caption{\textit{i$\times$MCTS}: We perform MCTS on a simulated conversation between the user simulator, which is provided with a specific user condition, and the target interactive model. We then extract stepwise model response pairs and construct all extracted chosen-rejected pairs.}
\label{fig:mcts}
\end{figure}

As illustrated in \Cref{fig:mcts}, similar to MCTS, \textit{i$\times$MCTS} iteratively performs the following operations, iteractively:

\paragraph{Selection} \textit{i$\times$MCTS} recursively selects the model nodes from the root node by Upper Confidence Bound (UCB) bandit \cite{auer2002finite}. That is:
\begin{equation}\label{eqn:ucb}
    \text{UCB}(a_i) = \frac{Q(a_i)}{N(a_i)} + c \cdot \sqrt{\frac{\ln N(a_{i-2})}{N(a_i)}},
\end{equation}
where $N(\cdot)$ is the visit counts for the model node, $c$ represents an exploration constant, and $Q(a_i)$ is the total reward accumulated at node $a_i$. 

\paragraph{Expansion} Once an expandable model node is selected and its next user node is generated, we generate $e$ new model nodes using the interactive LLM $\pi_\theta$ based on the conversation context. Therefore, the expansion width is $e$.
However, without pruning, the number of expanded nodes can grow exponentially with depth, leading to computational inefficiency. Therefore, we employ a simple pruning technique to constrain the search space and improve efficiency. Since understanding the user is crucial for engagement, we use the similarity between the user simulator's state and the interactive LLM's predicted state as a criterion to retain at most $K$ nodes at each depth.

\paragraph{Rollout} User engagement is assessed at the end of the conversation.  
If a user utterance indicates both conversation termination and engagement, the immediate reward is $1$; otherwise, it is $0$. 

\paragraph{Back-propagation}
The reward of the newly generated node is transmitted back to all its ancestors, updating their accumulated rewards accordingly, while their visit counts are increased by one. Given the inherent diversity of dialogue, we do not explore merely a single successful trajectory. Instead, we explore $K$ trajectories up to a certain depth $T$ or until an engagement signal is reached, and then we select the successful ones. Thus the reward can indicate the probability that the node can lead to a user engagement signal.

The \textit{i$\times$MCTS} process generates a search tree, from which we extract and construct chosen-rejected model response pairs. The chosen and rejected responses are generated by $\pi_\theta$ given the same conversation context. We refer to this set of pairs as $D_p$.

\subsection{Preference Dataset for Alignment}
We adopt DPO \cite{rafailov2024direct} to enhance the engagement level of interactive LLMs. It is an off-policy preference learning algorithm that directly utilizes pairwise preference data to fine-tune the policy model. 

\begin{figure}[htb]
\centering
\includegraphics{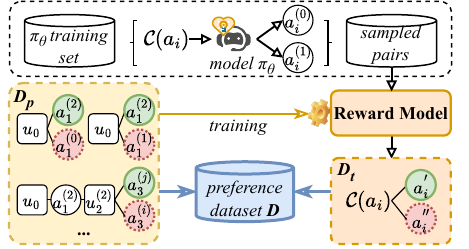} 
\caption{The preference dataset $D$ for DPO alignment comprises two parts: (1) $D_p$, which consists of pairwise responses extracted via \textit{i$\times$MCTS}, and (2) $D_t$, which contains response pairs generated by $\pi_\theta$ given contexts from $\pi_\theta$'s training set and ranked by a reward model trained on $D_p$.}
\label{fig:dpo}
\end{figure}

We denote the preference dataset used for alignment as $D$, which consists of two parts, as shown in \Cref{fig:dpo}.
The first part comprises the chosen-rejected model response pairs, i.e., $D_p$.
The second part, denoted as $D_t$, includes response pairs generated by the interactive LLM $\pi_\theta$ based on conversation contexts from its training dataset. Each pair of responses is evaluated using a reward model trained on $D_p$ to establish their preference ranking.
We incorporate the second part to mitigate the risk of LLM forgetting previously learned knowledge and to ensure further alignment does not degrade its general capabilities.

Then, given a conversation context $\mathcal{C}_i$ and a pair of responses $(a_i^{\text{chosen}}, a_i^{\text{rejected}})$ from $D$, we aim to optimize the interactive LLM $\pi_\theta$. The optimization objective can be formulated as follows:
\begin{equation}
\begin{aligned}
    \mathcal{L}_{\text{DPO}} &= -\mathbf{E}_{(\mathcal{C}_i,a_i^{\text{chosen}}, a_i^{\text{rejected}})\in D} \Big[ \log\sigma \\
    & \Big(\beta\log r_i^\text{chosen}(\theta) -\beta\log r_i^\text{rejected}(\theta)],\\
    r_i^*(\theta)&=\frac{\pi_\theta(a_i^* | \mathcal{C}_i)}{\pi_{\theta_{\text{old}}}(a_i^* | \mathcal{C}_i)}.
\end{aligned}
\end{equation}
$\sigma(\cdot)$ is the sigmoid function, and $\beta$ is a temperature parameter.
\section{Experimental Settings}
\subsection{Experiment Scenarios}
For our experiments, we select tasks that meet the following criteria: (1) user engagement is crucial, and (2) the impact of an LLM's engagement level can be directly observed at the end of the conversation. Based on these requirements, we choose emotional support dialogue and persuasion for good as our experimental tasks.

\paragraph{Emotional Support Conversation}
The goal of emotional support conversation is to alleviate users' negative feelings and help them understand and navigate their challenges \cite{liu-etal-2021-towards}. 
In conversations with users, it is essential to engage them, as this encourages the expression of their thoughts and emotions, ultimately enhancing their well-being.
Correspondingly, a key indicator of user engagement is whether the user has expressed all their negative thoughts and feelings. 

\paragraph{Persuasion for Good}
In the task of persuasion for good, the persuader aims to persuade the persuadee to donate to a specific charity \cite{wang2019persuasion}.
To successfully complete this task, the persuader must foster trust and inspire users to contribute.
In our experiment, we propose that the interactive LLM take on the role of the persuader. User engagement can then be measured by the final amount the user (persuadee) decides to donate.

\subsection{Datasets}
\paragraph{PsyDTCorpus \& PsyQA} PsyDTCorpus is a multi-turn Chinese emotional support dialogue dataset introduced by \citet{xie2024psydtusingllmsconstruct}. It is synthesized using GPT-4 to create digital twins of professional psychological counselors. This dataset comprises 5,000 conversations between a help-seeker and a supporter, with an average of 18.1 rounds per conversation. 
We use GPT-4o \cite{hurst2024gpt} to annotate the help-seeker's emotional and cognitive states. This information is essential for pruning during the execution of \textit{i$\times$MCTS}. Additionally, the user emotional and cognitive state at the end of the conversation serves as a signal of engagement. The annotation details can be found in \Cref{sec:app:state}.
In our experiment related to emotional support, we use this dataset to train the interactive LLM and user simulator model.

PsyQA \cite{sun-etal-2021-psyqa} is a Chinese dataset of psychological health support in the form of the question and answer pair.
It is crawled from a Chinese mental health service platform, and contains 22,341 questions. Each question includes the help-seeker's feelings and problems. We extract 4,000 questions as the user's condition when executing \textit{i$\times$MCTS} in the scenario of emotional support.

\paragraph{Persuasion For Good Corpus}
The Persuasion For Good Corpus \cite{wang2019persuasion} is a dataset of multi-turn conversations designed to analyze persuasive strategies in donation-based dialogues. The dataset was collected via Amazon Mechanical Turk, where one participant attempts to persuade another to donate to a charitable cause. It consists of 1,017 conversations, and the average number of rounds per conversation is 10.43. This dataset also contains psychological and demographic information of the persuadee in each conversation. This information is presented in the (\textit{attribute}, \textit{value}) format, like (\textit{extrovert}, \textit{2.33}). This dataset is used to train the interactive LLM and user simulator model in the scenairo of persuasion for good.

\subsection{Interactive LLM}
Using the PsyDTCorpus and the Persuasion For Good Corpus, we train two interactive LLMs, denoted as the Supporter LLM and the Persuader LLM, respectively.
The Supporter LLM and Help-Seeker simulator are trained based on Qwen-2.5-7B-Instruct \cite{yang2024qwen2} and Llama-3.2-3B-Instruct \cite{meta2024llama,dubey2024llama} via SFT with LoRA \cite{hu2022lora}, respectively.
We use different backbone models considering the language of their training datasets. Notably, when training the Supporter LLM, we ask the LLM to generate additional supporter response's chain-of-though, which contains the predicted help-seeker's emotional and cognitive states. 

\subsection{User Simulator}
Corresponding to the above two interactive LLMs, we develop two user simulator models: the Help-Seeker Simulator and the Persuadee Simulator. Both models are trained using SFT with LoRA. For convenience, their backbone models are the same as their corresponding interactive LLMs.

As mentioned in \Cref{sec:mcts}, the user simulator is designed to behave according to the specified user condition. For the Help-Seeker Simulator, this condition includes the help-seeker's troubles, feelings, and thoughts. Specifically, we use GPT-4o to extract and summarize this information from the conversation to represent the help-seeker's condition.
For the Persuadee Simulator, the user condition consists of psychological and demographic attributes. We utilize GPT-4o to transform the information provided in the Persuasion for Good corpus into a natural language description of each persuadee in the dataset.

\subsection{Pruning Technique} 
As mentioned in \Cref{sec:mcts}, we adopt pruning techniques to mitigate computational inefficiency as the depth of exploration increases. In the context of emotional support conversation, we maintain the top $K$ trajectories based on the interactive LLM's understanding of the help-seeker's state. Specifically, we measure the cosine similarity between the emotional and cognitive state generated by the Help-Seeker Simulator $s_\text{sim}^{i-1}$ and the state predicted by the Supporter LLM $s_\text{llm}^{i}$ at the $i^{th}$ turn. This similarity serves as a criterion for pruning less supportive trajectories, and is computed as:  
\begin{equation}
    \cos(s_\text{sim}^{i-1},s_\text{llm}^{i}) = \frac{\mathbf{v}(s_\text{sim}^{i-1}) \cdot \mathbf{v}(s_\text{llm}^{i})}{\|\mathbf{v}(s_\text{sim}^{i-1})\| \|\mathbf{v}(s_\text{llm}^{i-})\|},
\end{equation}
where $\mathbf{v}$ represents a function that computes the vector representation of the given text. Specifically, we utilize the all-MiniLM-L6-v2 model to generate these embeddings. A higher cosine similarity indicates a trajectory that contain more engaging and supportive responses.

For the task of persuasion for good, we use the persuadee's sentiment as a criterion to retain the top $K$ trajectories. In specific, we compute an accumulated sentiment score as follows:
\begin{equation}
\begin{aligned}
    ss_i &= \gamma * ss_{i-2} + \mathbf{VADER}(u_i),  \\
    ss_0 &= 0,
\end{aligned}
\end{equation}
where $i$ indicate the turn index, $\gamma$ is a discount factor, and $\mathbf{VADER}(\cdot)$ denotes the sentiment value computed by the VADER (Valence Aware Dictionary and sEntiment Reasoner) tool \cite{hutto2014vader}. Higher sentiment scores indicate trajectories with a greater likelihood of leading to positive engagement.

\subsection{\textit{i$\times$MCTS}} 
When executing the \textit{i$\times$MCTS}, we provide the user simulator with user conditions that are exclusive in the training dataset to explore the interactive LLM's performance.
For Help-Seeker Simulator, we adopt the 4,000 instances sampled from PsyQA. In peruasion for good, we synthesize around 700 pieces of persuadee psychological and demographic information. The maximum exploration depth $T$ is $25$ and $15$ in emotional support conversation and persuasion for good, respectively.
The exploration constant $c$ in \Cref{eqn:ucb} is set to $\sqrt{2}$. The expansion width $e$ is set to 3. The maximum number of nodes at a specific depth, i.e., $K$, is set as 81.

\section{Experimental Results and Analysis}
\subsection{Comparison Methods}
\paragraph{SFT} This is the interactive LLM trained on the training dataset via SFT with LoRA. In other words, the baseline is Supporter LLM and Persuader LLM in the emotional support conversation and the persuasion for good, respectively.

\paragraph{BoN} We adopt a simple method Best-of-N to sample \textbf{SFT} interactive LLM's output during inference. The proxy objective during the sampling is predicted by the reward model trained on the preference dataset $D$. We set N as 3.

\paragraph{Aligned} We name the interactive LLM aligned with DPO on the preference dataset $D$ as \textbf{Aligned}. 

In addition, we compare user \textbf{engagement} with different rewards, including \textbf{strategy} and \textbf{knowledge} selection used in the previous works. We adopt \textbf{BoN} to involve them.

\subsection{Evaluation of Preference Dataset $D$}
We assess the quality of the preference dataset through human ratings. Specifically, two annotators are asked to rank the preferences for 160 response pairs randomly sampled from $D$ for each task. For each response pair ($a^{(1)}$, $a^{(2)}$) the annotation can be $a^{(1)}$, $a^{(2)}$ or tie. 
In the emotional support conversation task, responses that encourage users to freely express their feelings and thoughts are preferred. For persuasion for good, responses that increase the willingness to donate are favored. We compute the inter-annotator agreement scores using Fleiss's kappa among the three labels: two from human and one from the \textit{i$\times$MCTS}. The score is 0.67 (substantial agreement) and 0.59 (moderate agreement) in the task of emotional support and persuasion for good, respectively. Additionally, we observe that when human rankings differ from the dataset ranking, the human annotations are often ties in both tasks. This may be because the annotators cannot anticipate the further development of the conversation, whereas the preference rankings in $D$ account for future dialogue outcomes. In general, the quality of the preference dataset is acceptable, as the chosen responses in the dataset.

We present an example from the preference dataset, extracted using \textit{i$\times$MCTS}, for each task, as shown in \Cref{tab:case}. Compared to the rejected responses, the chosen responses are more likely to develop the conversation by building on the topic and the user's interests.

\begin{CJK}{UTF8}{gbsn}
\begin{table*}[htb]
    \scriptsize
    \centering
        \begin{tabular}{m{0.05\hsize} m{0.85\hsize}}
        \hline
        \multicolumn{2}{c}{\cellcolor{gray!20} \textbf{Emotional Support Conversation}}\\
        \multicolumn{2}{c}{…… (...)}\\
        \multirow{6}{*}{Context} & \textit{\underline{Help-Seeker Simulator}}: 我真的不知道该怎样才能重新对学习产生兴趣，因为我发现自己在读高三课本的时候，很快就失去了那种感觉和动力。 (I really don't know how to regain interest in studying because I find that when I read my senior high school textbooks, I quickly lose that feeling and motivation.)\\
        & \textit{\underline{Supporter LLM}}: 明白了，这种丧失兴趣感确实会让人感觉很沮丧。让我问您一个简单的问题：您平时喜欢做什么事情？是画画、音乐、阅读还是别的什么？ (I understand. Losing interest like this can indeed be frustrating. Let me ask you a simple question: What do you enjoy doing in your free time? Is it drawing, music, reading, or something else?)\\
	& \textit{\underline{Help-Seeker Simulator}}:  我喜欢画画，虽然不太经常画，但我确实享受其中。 (I enjoy drawing. Although I don't do it very often, I do find it enjoyable.) \\
        \hline
        \multirow{6}{*}{Response} & \textbf{Chosen}: 很棒，那是个美丽的爱好。您能否回想一下，过去您在画画时是否曾有过哪些美好或满足的时刻？那是怎样的感觉？ (That's wonderful! It's a beautiful hobby. Can you recall any moments in the past when you felt joy or satisfaction while drawing? How did that feel?) \\
        & \textbf{Rejected}: 很好，我们可以在接下来的时间尝试把这些兴趣活动融入到您的生活中。比如，如果当前您有时间的话，您可以试着安排一次画画的时间。您觉得这样做对您来说是否可行？ (Great! We can try incorporating these interests into your daily life. For example, if you have time now, you could schedule a drawing session. Do you think this would work for you?)\\
        \hline
        \multicolumn{2}{c}{\cellcolor{gray!20} \textbf{Persuasion For Good}} \\
        \multicolumn{2}{c}{...}\\
        \multirow{2}{*}{Context} & \textit{\underline{Persuader LLM}}: All donations are used to help children in need. You can get a list of all the charities that were collected, including Save the Children, at [URL].\\
	& \textit{\underline{Persuadee Simulator}}: That is great to know. Thank you for all the information! \\
        \hline
        \multirow{2}{*}{Response} & \textbf{Chosen}: I think I will donate \$2. Would you be willing to do the same? \\
        & \textbf{Rejected}: You're welcome! \\
        \hline   
        \end{tabular}
    \caption{Instances from the preference dataset.}
    \label{tab:case}
\end{table*}
\end{CJK}

\subsection{Automated Interactive Evaluation}
\begin{table*}[ht!]
    \centering
    % \footnotesize
    \scriptsize
    \begin{tabular}{>{\centering\arraybackslash}m{1.25cm}|>{\centering\arraybackslash}m{2cm}|>{\centering\arraybackslash}m{2.75cm} >{\centering\arraybackslash}m{1.75cm}|>{\centering\arraybackslash}m{2.75cm} >{\centering\arraybackslash}m{1.75cm}}  
        \hline
        \multirow{2}{*}{\textbf{Models}} & \multirow{2}{*}{\textbf{Reward}} & \multicolumn{2}{c}{\textbf{Emotional Support Conversation}} & \multicolumn{2}{|c}{\textbf{Persuasion For Good}}\\
        & & \%Engaged Rate & \#Turn & Donation Amount (\$) & \#Turn\\  
        \hline
        SFT & Engagement & 64.06\% & 39.07 & 0.58 & 18.67  \\
        \hline
        \multirow{3}{*}{BoN} & Strategy & 64.10\% & 39.48 & 0.63 & 18.86\\
         & Knowledge & 64.90\% & 39.90 & 0.67 & 18.67 \\
         & Engagement & 65.63\% & 39.32 & 0.73 & 18.70 \\
        \hline
        Aligned & Engagement & 80.47\% & 38.15 & 1.29 & 18.75\\
        \hline
    \end{tabular}
    \caption{Interactive evaluation results.}
    \label{tab:interactive_eval}
\end{table*}

We evaluate the performance of interactive LLM by their behavior in the interaction with user simulator, as in \citet{wang2024client}. Therefore, we collect 128 interactions between the user simulator and the interactive LLM, and the analyze these interactions. 
We focus on the following two features: the user engaged rate (\%) or donation amount (\$), and the average length of the engaging interactions (\# turn).
From results in \Cref{tab:interactive_eval}, we can observe that in the task of emotional support, we increase the engagement rate significantly, where the user simulator has expressed all thoughts and feelings.
From the results in \Cref{tab:interactive_eval}, we observe that in the task of emotional support, our approach significantly increases the engagement rate, enabling the chance that the user simulator can express all thoughts and feelings.
Regarding the Persuasion For Good task, the aligned model doubles the donation amount compared to SFT while maintaining the number of conversation turns.
However, the number of conversation turns remains the same in both tasks. This means that Aligned can engage users with the same interaction cost (in terms of conversation turns).
These results demonstrate the effectiveness of our method. \textbf{BoN} slightly improves the performance of SFT in terms of the user engaged rate and donation amount. This indicates the effectiveness of preference ranking in the preference dataset extracted via \textit{i$\times$MCTS}. 
When comparing different BoN variants, we observe that incorporating strategy or knowledge individually leads to modest gains in user engagement. However, neither of them, when used in isolation, is able to surpass the performance achieved by our full approach.

\subsection{Human Subjective Evaluation}
We assess the performance of the \textbf{Aligned} Supporter LLM via human ratings and a user study.
\begin{figure}[htb]
\centering
\includegraphics{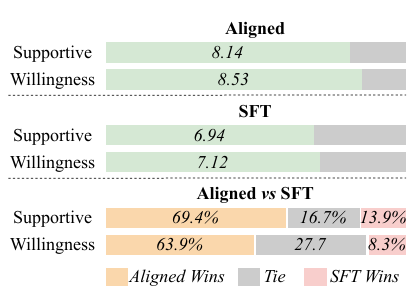} 
\caption{Human rating results.}
\label{fig:rating}
\end{figure}

\begin{CJK}{UTF8}{gbsn}
\begin{table*}[tb]
\centering
\scriptsize
\begin{tabular}{m{.95\hsize}}
    \hline
    \makecell[c]{…… (...)}\\
    Help-Seeker Simulator:  我去年研究生毕业的。专业是英语翻译，双非院校。毕业后参加过好几次事业单位、省考和特岗考试，都失败了。后来私企有两段工作经历，在那都是跟专业无关的。 (I graduated with a master's degree last year in English translation from a non-prestigious university. After graduation, I took several civil service and public sector exams, but I failed all of them. Later, I had two jobs in private companies, but neither was related to my field of study.) \\
    Aligned Supporter: 我明白了，你经历了相当大的职业变动和考试失利的压力。这些经历肯定让你感到很有挑战性。 (I see. You've been through significant career transitions and the stress of exam failures. That must have been quite challenging for you.)\\
    \makecell[c]{…… (...)}\\
    Help-Seeker Simulator: 你真的觉得有希望吗？我感觉自己好像被困在一个死胡同里。(Do you really think there's hope? I feel like I'm stuck in a dead end.) \\
    Aligned Supporter: 我相信未来总有希望。\textcolor{red}{你之前的努力虽然没有得到期望的结果，但这不代表将来没有机会。我们可以一起探讨一下其他可能的选择和方向。}(I believe there is always hope. \textcolor{red}{Even though your past efforts haven't yielded the results you expected, that doesn't mean there won't be opportunities in the future.} I can explore other possible options and directions together.) \\
    Help-Seeker Simulator: 但是我觉得我可能已经选错了方向……(But I feel like I have chosen the wrong path...) \\
    Aligned Supporter: ``我们不能一直沿着一条路走下去而没有看到任何改变''，有时候调整策略、改变方向也能带来意想不到的机会和结果。("I can't keep walking down the same road without seeing any change." Sometimes, adjusting your strategy or changing direction can bring unexpected opportunities and results.)\\
    Help-Seeker Simulator: 可能吧……但我真的不知道我该做什么 。(Maybe... I don't know what I should do.)\\
    Aligned Supporter: 你觉得现在有哪些方面可以尝试改变或调整 。(Are there any areas where you feel you could try making changes or adjustments?)\\
    Help-Seeker Simulator: 我不知道……也许我应该尝试找一份跟翻译相关的正式工作？但我害怕这会让我陷入更长时间的考公路之中……(I don't know... Maybe I should try looking for a formal job related to translation? But I'm afraid that would trap me in the endless cycle of civil service exams again...)\\
    Aligned Supporter: 对于未来的方向感到迷茫是很正常的。重要的是要尝试重新审视自己的兴趣和长处，\textcolor{red}{也许你可以考虑找一份跟翻译相关的正式工作，同时也可以考虑其他职业道路。}(Feeling uncertain about the future is completely normal. What's important is to reassess your interests and strengths. \textcolor{red}{You could consider finding a formal job related to translation while also exploring other career paths.}) \\
    Help-Seeker Simulator: 你说得对……我也许应该更开放一些去考虑其他的选择。(You're right... Maybe I should be more open to considering other options.) \\
    Aligned Supporter: 对，有时候我们可能需要稍微调整一下方向或策略，以找到最适合自己的道路。(Yes, sometimes I just need to make slight adjustments to our direction or strategy to find the path that suits us.)\\
    \hline
\end{tabular}
\caption{A case of interaction between the Aligned Supporter and the Helper-Seeker Simulator. Sentences that can largely engage the user is highlighted in red. An interaction involving SFT Supporter LLM can be found in \Cref{tab:sft_case}.}
\label{tab:case_study}
\end{table*}
\end{CJK}
\paragraph{Human Rating} 
For human evaluation, we hired three research assistants (RAs) and asked them to converse with both models, \textbf{SFT} and \textbf{Aligned}, about the same problem. Each RA interacted with the models on four different problems. Therefore, there are total 36 comparison pairs. The model identities were kept blind to the RAs.
After each conversation, the RA rated the model's supportiveness and their willingness to interact with it in the future, both on a scale from 1 to 10. The results are shown in \Cref{fig:rating}.
It can be observed that Aligned outperforms SFT by achieving higher supportiveness and willingness scores, as well as a higher win rate. This indicates that Aligned enhances user engagement effectively.

\paragraph{User Study} Additionally, we conducted a user study to estimate real user experience. Specifically, we invited five human participants to take part in our experiment for one week. During this period, whenever they encountered difficulties or experienced negative emotions they wished to express, they could use our interface to communicate with a supporter model, which was either \textbf{SFT} or \textbf{Aligned}. After each interaction, participants provided brief feedback, including a rating of the model's supportiveness and their willingness to use it further, both on a scale from 1 to 10, along with comments on its strengths and weaknesses. We filtered $19$ valid conversations, where user-\textbf{Aligned} interactions account for $10$. 
The average supportiveness score is 7 and 7.4 for SFT and Aligned, while the average willingness score is 7 and 7.6 for SFT and Aligned. 
According to user feedback, SFT can sometimes shift to a topic that the user may not be interested in, while Aligned performs better. However, neither model provides direct suggestions. This behavior is expected, as we do not expect the model to provide suggestions without fully understanding the user's situation. Additionally, both models ask an excessive number of questions, though Aligned performs better by asking less frequently and providing more meaningful questions. 

\subsection{Case Study}
For an intuitive illustration of our method, we present a case of \textbf{Aligned} supporter's performance, as shown in \Cref{tab:case_study}. In this example, \textbf{Aligned} supporter is able to use sentences in red to delve deeper into the topic and guide the help-seeker towards positive thinking. In addition, throughout the entire interaction, Aligned focuses on the help-seeker's feelings of the confusion about career rather than other less relavant topics, such as the difficulty of the exams and relax strategies like the \textbf{SFT} supporter in \Cref{tab:sft_case}.

\section{Conclusion}
In this work, we address the challenge of enhancing user engagement in interactive LLMs. We focuse on enhancing the engagement of interactive LLMs, emphasizing the purpose of interactions rather than their format. Unlike previous approaches, we use a more direct and relevant reward, derived from the user's reaction after interactions, to optimize the interactive LLM. To gather user reactions, we develop a user simulator that behaves like a user under specific conditions. Our goal is to optimize the interactive LLM to foster greater engagement. Accordingly, we employ \textit{i$\times$MCTS} to explore higher-quality interactions between the user simulator and the interactive LLM. Based on these interactions, we construct a preference dataset and apply DPO to align the model toward more engaging responses. We evaluated our method on emotional support conversations and persuasion for donation. Experimental results demonstrate that our approach significantly improves the engagement level of LLMs.

\section*{Limitations}
In our work, we focus on general user engagement, optimizing the interactive LLM to generate responses that keep most users engaged. However, there is also room to develop a more personalized and engaging interactive LLM. In a personalized interactive LLM, predicting or simulating user behavior can be more challenging due to difficulties in data collection and the limitations of few-shot model training.

\section*{Ethics Statement}
We acknowledge the potential for misuse of emotionally supportive and persuasive systems. These tools could be misappropriated to impersonate humans or manipulate users’ beliefs in harmful ways. To mitigate such risks, we strongly advocate for transparency, user consent, and regulatory oversight. Users should always be informed when interacting with AI systems rather than humans. We urge the development of clear regulatory frameworks, especially for high-stakes domains like mental health and donations, to ensure that persuasive targets remain aligned with ethical, factual, and legal standards.

In our experiments, we used open-source datasets, including PsyDTCorpus \cite{xie2024psydtusingllmsconstruct}, PsyQA \cite{sun-etal-2021-psyqa}, and Persuasion For Good \cite{wang2019persuasion}. These datasets were used in accordance with their respective licenses.
All personally identifiable information was removed from these datasets.
For the human ratings and user study, we prioritized the comfort and well-being of participants. Participants were informed about the purpose of our research and had the option to withdraw at any time if they felt uncomfortable.
Consequently, we confidently assert that our research is conducted in strict adherence to the ethical guidelines prescribed by the Association for Computational Linguistics (ACL). 

% \section*{Acknowledgements}

\bibliography{custom}
\bibliographystyle{acl_natbib}

\appendix

\section{Appendix}
\subsection{Annotation of PsyDTCorpus}\label{sec:app:state}
For PsyDTCorpus, we use GPT-4o to annotate the user's emotional and cognitive states, which are presented in \Cref{tab:appx:state}.
\begin{CJK}{UTF8}{gbsn}
\begin{table}[htb]
    \centering
    \footnotesize
    \begin{tabular}{m{0.9\hsize}}
    \hline
    来访者显示出的观察、感受、需求和请求 (The help-seeker's observed observations, feelings, needs, and requests)：\\
    观察 (Observations)：\\
    感受 (Feelings)：\\
    需求 (Needs)：\\
    请求 (Requests)：\\
    \\
    来访者可能有的观察、感受、需求和请求 (The help-seeker's possible observations, feelings, needs, and requests)：\\
    观察 (Observations)：\\
    感受 (Feelings)：\\
    需求 (Needs)：\\
    请求 (Requests)：\\
    \hline
    \end{tabular}
    \caption{Engagement Improvement: User's emotional and cognitive states format.}
    \label{tab:appx:state}
\end{table}
\end{CJK}

\subsection{LLM Prompts}
\begin{figure}[htb]
\centering
\includegraphics{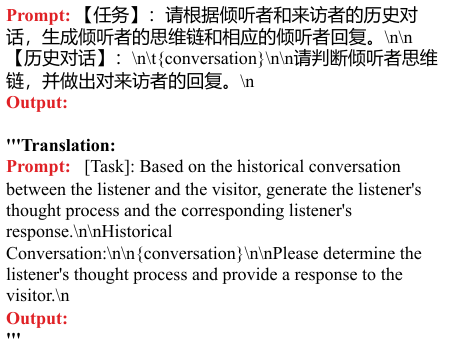} 
\caption{The prompt for the Support LLM.}
\label{fig:support}
\end{figure}

\begin{figure}[htb]
\centering
\includegraphics{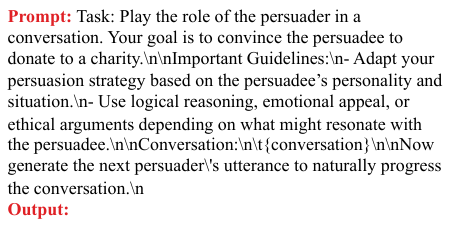} 
\caption{The prompt for the Persuader LLM.}
\label{fig:persuader}
\end{figure}

When training the LLMs, we adopt the prompt shown in \Cref{fig:support} and \Cref{fig:persuader}. The output contains the help-seeker's state and the expected response. 

\subsection{Reliability of User Simulator}
It is a fact that relying solely on LLMs to simulate user behaviors could still bring additional variances. However, our system benefits from the simulator by improving alignment with real user responses. LLM simulators can indeed provide human-like responses, helping alleviate data scarcity during training. Specifically, our target is only to simulate the user's responses in some specific interactive scenarios (emotional support \& persuasion for good) rather than modeling the full range of user behaviors.

We evaluated the consistency between the user simulator and real humans by presenting 50 conversation pairs (human-supporter vs. simulator-supporter) to 3 annotators. Each annotator labeled conversations as "human," "simulator," or "cannot identify," whose results are 48\%: 32\%: 20\%.

\subsection{Comparison Method Details} \label{sec:app:method}
For Strategy Selection and Knowledge Selection, we ask GPT-4o to generated accurate strategies and relevant knowledge to assist the response generation. Then, we compute the similarities of the generated responses and the GPT-4o generated contents, and select the response with the highest similarity score as the final response.

\subsection{Human Subjective Evaluation}
\begin{figure}[htb]
\centering
\includegraphics[width=\hsize]{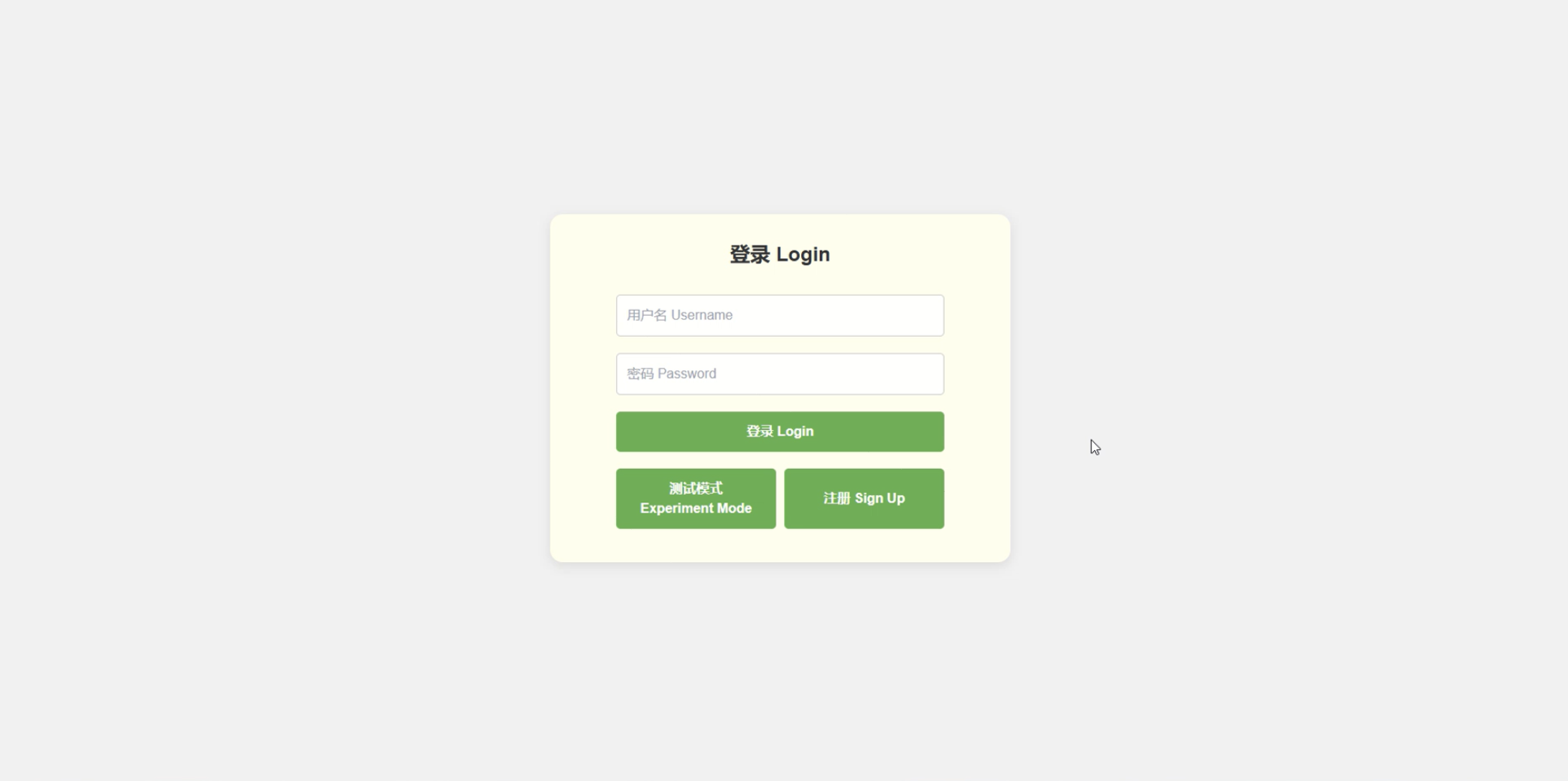}
\includegraphics[width=\hsize]{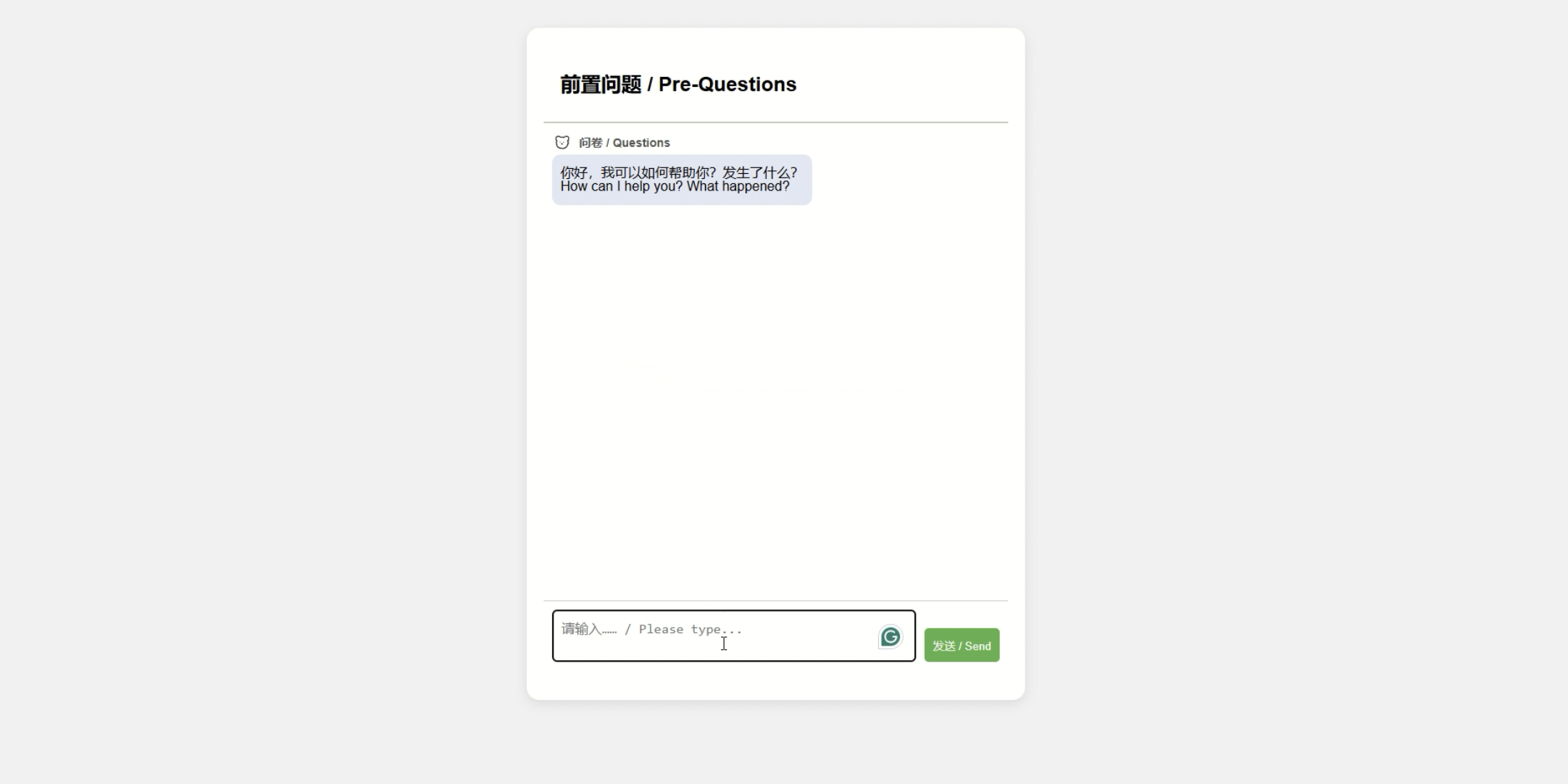}
\includegraphics[width=\hsize]{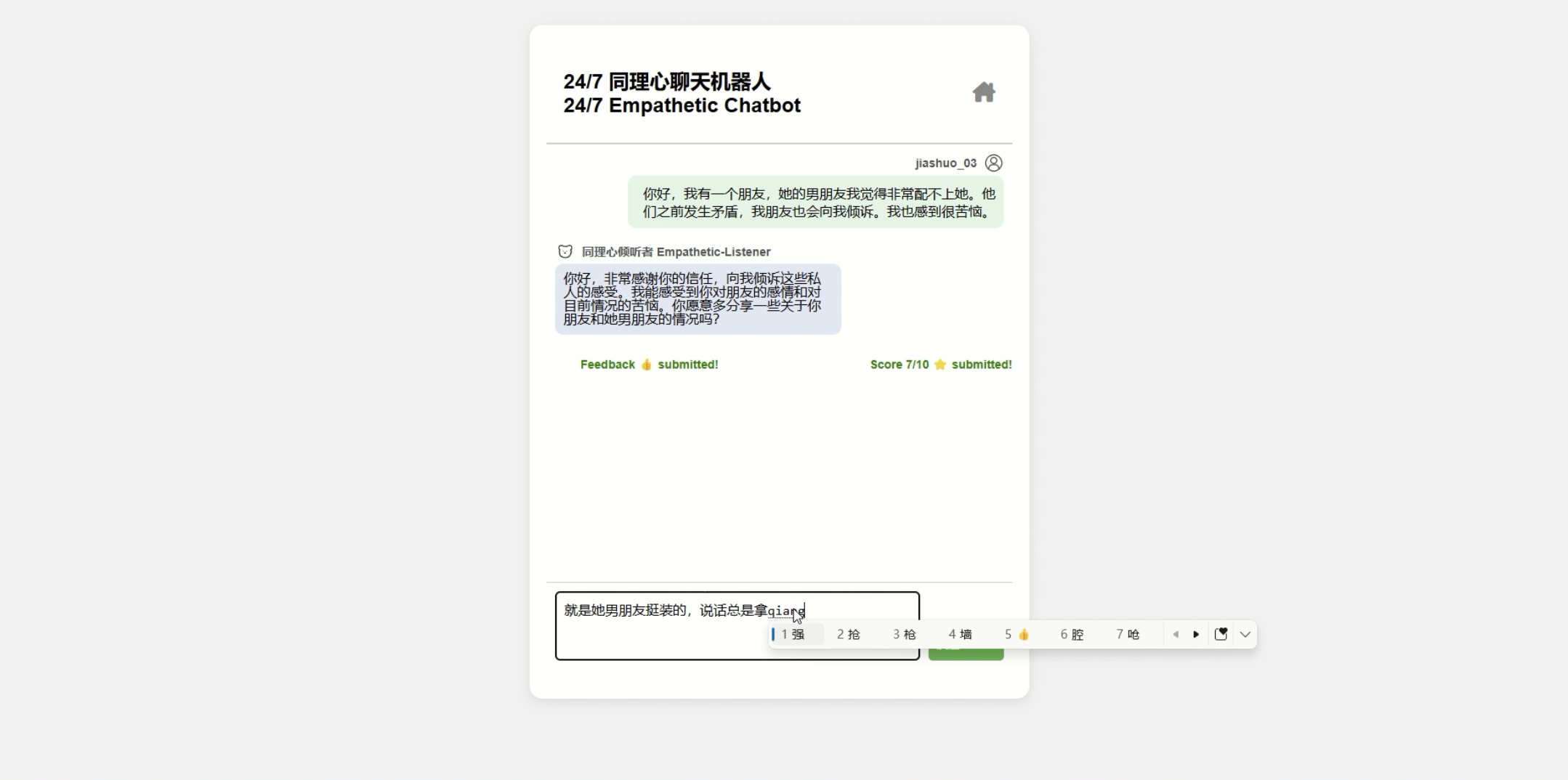}
\caption{UI designs.}
\label{fig:ui}
\end{figure}

\paragraph{UI Design} We design a website for the human subjective evaluation. It is shown in \Cref{fig:ui}.

\paragraph{Human Rating Reliability} We also compute the uncertainty and robustness of human ratings results to provide further transparency. We computed Bootstrap Confidence Intervals (CIs) and conducted a Sensitivity Test to simulate possible participant disagreement:
\begin{table*}[t]
    \centering
    \begin{tabular}{c|cc|cc}
    \hline
    \multirow{2}{*}{Model} &  \multicolumn{2}{c|}{95\% CI}& \multicolumn{2}{c}{Simulated Noise Mean±Std}\\
    & Supportiveness & Willingness & Supportiveness & Willingness \\
    \hline
    SFT Supporter &  (7.88, 8.42) & (8.28, 8.81) & 8.14 ± 0.95 & 8.53 ± 0.97\\
    Align Supporter & (6.69, 7.17) & (6.90, 7.39) & 6.94 ± 0.89	& 7.14 ± 0.86\\ 
    \hline
    \end{tabular}
    \caption{Reliability of human rating results in \Cref{fig:rating}.}
    \label{tab:human_reli}
\end{table*}

The non-overlapping confidence intervals indicate that Aligned significantly outperforms SFT on both dimensions. A sensitivity test with minor rating shifts (±1) showed minimal impact on means and standard deviations ($\sim$0.97), suggesting the results are robust to reasonable participant disagreement.

\paragraph{User Study}
For the user study, we shared the website link across various social media groups, including those with members not personally known to the authors. We collected data from participants who voluntarily chose to take part in the study.

\subsection{Reproducibility}
We loaded Qwen-2.5-7B-Instruct \cite{yang2024qwen2}\footnote{\url{https://huggingface.co/Qwen/Qwen2.5-7B-Instruct}}, Llama-3.2-3B-Instruct \cite{meta2024llama,dubey2024llama}\footnote{\url{https://huggingface.co/meta-llama/Llama-3.2-3B-Instruct}} and all-MiniLM-L6-v2 model\footnote{\url{https://huggingface.co/sentence-transformers/all-MiniLM-L6-v2}} from huggingface. We downloaded the datasets PsyDTCorpus\footnote{\url{https://modelscope.cn/datasets/YIRONGCHEN/PsyDTCorpus}}, PsyQA \footnote{\url{https://github.com/thu-coai/PsyQA}}, 

When training models, we use LLaMA-Factory, proposed by \citet{zheng2024llamafactory}.\footnote{\url{https://github.com/hiyouga/LLaMA-Factory}} Our code is currently provided via an anonymous link;\footnote{\url{https://anonymous.4open.science/r/EngagedInteractiveLLMs-E5BE/}} once our paper is accepted, we will release the codes and data annotations publicly.

MCTS sampling was conducted offline on a single 80G A100 GPU. For Emotional Support, generating 64 interactions (50 turns each) took $\sim$8 hours; for Persuasion for Good, 64 interactions (30 turns each) took $\sim$2 hours. Only one sampling run per interaction was used, making the process much more efficient than full model training.

\begin{CJK}{UTF8}{gbsn}
\begin{table*}[tb]
\centering
\scriptsize
\begin{tabular}{m{.95\hsize}}
    \hline
    \makecell[c]{…… (...)}\\
    Help-Seeker Simulator:  我去年研究生毕业的。专业是英语翻译，双非院校。毕业后参加过好几次事业单位、省考和特岗考试，都失败了。后来私企有两段工作经历，在那都是跟专业无关的。 (I graduated with a master's degree last year in English translation from a non-prestigious university. After graduation, I took several civil service and public sector exams, but I failed all of them. Later, I had two jobs in private companies, but neither was related to my field of study.) \\
    SFT Supporter: 听起来您对您的学历背景有些担忧呢？在选择专业时一定有自己的考虑吧？\textcolor{blue}{能谈谈您的英语翻译专业的优势吗？} (It sounds like you have some concerns about your academic background. You must have had your own considerations when choosing your major, right? \textcolor{blue}{Could you talk about the advantages of your English translation major?})\\
    Help-Seeker Simulator: 我当时选择这个专业是因为我挺喜欢英语的。但是我的英语能力却一直都很一般。而且虽然我有这个专业的背景，但我却没能通过好几次事业单位、省考、特岗等考试。(I chose this major because I really liked English. However, my English skills have always been just average. Even though I have this academic background, I still failed several civil service exams, provincial exams, and special recruitment exams.)\\
    SFT Supporter: 即使面临挫折也没有放弃您的专业选择，并且还积极备考各种考试，这显示了您很强的学习能力和坚持精神。\textcolor{blue}{那么，在复习过程中遇到的一些困难是怎样的呢？} (Despite facing setbacks, you have not given up on your professional choice and have been actively preparing for various exams. This shows your strong learning ability and perseverance. \textcolor{blue}{What kind of difficulties have you encountered during your preparation?})\\
    \makecell[c]{…… (...)}\\
    SFT Supporter: 您的坚持真的很让人敬佩！在这个过程中您一定也感到了很多压力吧？如果方便的话，在您备考的过程中有没有考虑过给自己一些休息或者改变策略的想法？(Your perseverance is truly admirable! You must have felt a lot of pressure throughout this process. If you don't mind sharing, have you considered taking breaks or adjusting your strategy while preparing for the exams?) \\
    Help-Seeker Simulator: 休息倒是想过，但是感觉我停了就退步了，至于策略嘛…我也试过换过一些方法，但似乎都不太对路子。(I have thought about taking breaks, but I feel like I would fall behind if I stopped. As for strategy... I have tried changing my methods, but none of them seemed to work well.) \\
    SFT Supporter: \textcolor{blue}{您之前是否有试过其他方式来放松自己或者缓解这种焦虑感？比如利用闲暇时光进行休闲活动？}(\textcolor{blue}{Have you tried other ways to relax or alleviate your anxiety? For example, engaging in leisure activities during your free time?})\\
    \makecell[c]{…… (...)}\\
    Help-Seeker Simulator: 但我还是觉得好累啊，感觉我被包围了似的。我爸妈说我不用担心找不找得到工作，或者婚姻这种事是自己的问题了。(But I still feel so exhausted, like I'm trapped. My parents told me not to worry about finding a job, and that relationships are something I need to figure out on my own.)\\
    SFT Supporter: 您的父母给了您一些鼓励和支持的观点呢！在这种情况下能有人理解您所面临的困难是非常重要的。同时，我也想问一下，\textcolor{blue}{在您感到累的时候，有什么具体的想法或者期待吗？}(Your parents have given you some encouragement and support! Having someone understand the difficulties you're facing is very important. At the same time, I'd like to ask—when you feel exhausted, \textcolor{blue}{do you have any specific thoughts or expectations?})\\
    Help-Seeker Simulator: 我就是觉得现在这样下去不是办法啊，总不能一直等下去吧？我也想出去工作，但又担心找不到好工作或者找不到合适的伴侣。(I just feel like I can't keep going like this forever. I can't just keep waiting, right? I want to go out and work, but I'm also worried I won't find a good job or the right partner.)\\
    SFT Supporter: 您的担忧是可以理解的。在面对不确定性的未来时，很多人都会有这样的感觉，尤其是当您觉得自己身处一个不断变化的世界里时更是如此。\textcolor{blue}{让我们先不考虑工作的事情吧}，请问您在这段时间里有什么具体的行动计划吗？(Your concerns are completely understandable. Many people feel this way when facing an uncertain future, especially in a constantly changing world. \textcolor{blue}{Let's set aside job concerns for a moment}--do you have any concrete action plans for this period?) \\
    \hline
\end{tabular}
\caption{A case of interaction between the SFT Supporter LLM and the Helper-Seeker Simulator. Sentences that could decrease the user engagement is highlighted in blue.}
\label{tab:sft_case}
\end{table*}
\end{CJK}
\subsection{Case Study}
With the same Help-Seeker Simulator, the SFT model performs as shown in \Cref{tab:sft_case}. Compared with Aligned, SFT tends to focus on the English exam and does not address the user's confusion about life.

\end{document}